\documentclass[letterpaper, 10 pt, conference]{ieeeconf}  

\IEEEoverridecommandlockouts                              

\usepackage{graphics} 
\usepackage{epsfig} 
\usepackage{mathptmx} 
\usepackage{times} 
\usepackage{amsmath} 
\usepackage{amssymb}  
\usepackage{marvosym}
\usepackage{subfig}
\PassOptionsToPackage{table}{xcolor}
\usepackage{xcolor}
\usepackage{colortbl}
\usepackage{booktabs}
\usepackage{caption}
\usepackage{cite}
\usepackage{hyperref}

\usepackage[nomargin,inline,draft]{fixme} 

\FXRegisterAuthor{yf}{ayf}{Yifan}

\definecolor{cvprblue}{rgb}{0.21,0.49,0.74}
\definecolor{forestgreen}{rgb}{0.0, 0.5, 0.0}
\definecolor{darkpastelgreen}{rgb}{0.01, 0.75, 0.24}
\definecolor{darkgreen}{rgb}{0.00, 0.8, 0.2}
\definecolor{darkyellow}{rgb}{0.96, 0.75, 0.00}

\title{\LARGE \bf
Delving into Mapping Uncertainty for Mapless Trajectory Prediction
}

\author{Zongzheng Zhang$^{1, 2*}$, Xuchong Qiu$^{2*}$, Boran Zhang$^{1}$, Guantian Zheng$^{1}$, Xunjiang Gu$^{4}$,\\ Guoxuan Chi$^{1}$,  Huan-ang Gao$^{1}$, Leichen Wang$^{2\dagger}$, Ziming Liu$^{2}$, Xinrun Li$^{2}$, \\Igor Gilitschenski$^{4}$, Hongyang Li$^{5}$, Hang Zhao$^{3}$, Hao Zhao$^{1}\textsuperscript{\Letter}$
\thanks{$^{1}$Institute for AI Industry Research (AIR), Tsinghua University, China.}
\thanks{$^{2}$Bosch Corporate Research, China.}
\thanks{$^{3}$Institute for Interdisciplinary Information Sciences(IIIS), Tsinghua 
              University, China.}
\thanks{$^{4}$University of Toronto, Canada.}
\thanks{$^{5}$University of HongKong, China.}
\thanks{$*$ Equal contribution.}
\thanks{$^{\dagger}$ Project lead.}
\thanks{\textsuperscript{\Letter} Corresponding to zhaohao@air.tsinghua.edu.cn}
}

\begin{document}

\maketitle
\thispagestyle{empty}
\pagestyle{empty}

\begin{abstract}
Recent advances in autonomous driving are moving towards mapless approaches, where High-Definition (HD) maps are generated online directly from sensor data, reducing the need for expensive labeling and maintenance. However, the reliability of these online-generated maps remains uncertain. While incorporating map uncertainty into downstream trajectory prediction tasks has shown potential for performance improvements, current strategies provide limited insights into the specific scenarios where this uncertainty is beneficial. In this work, we first analyze the driving scenarios in which mapping uncertainty has the greatest positive impact on trajectory prediction and identify a critical, previously overlooked factor: the agent’s kinematic state. Building on these insights, we propose a novel Proprioceptive Scenario Gating that adaptively integrates map uncertainty into trajectory prediction based on forecasts of the ego vehicle’s future kinematics. This lightweight, self-supervised approach enhances the synergy between online mapping and trajectory prediction, providing interpretability around where uncertainty is advantageous and outperforming previous integration methods. Additionally, we introduce a Covariance-based Map Uncertainty approach that better aligns with map geometry, further improving trajectory prediction. Extensive ablation studies confirm the effectiveness of our approach, achieving up to 23.6\% improvement in mapless trajectory prediction performance over the state-of-the-art method using the real-world nuScenes driving dataset. Our code, data, and models are publicly available at \href{https://github.com/Ethan-Zheng136/Map-Uncertainty-for-Trajectory-Prediction}{https://github.com/Ethan-Zheng136/Map-Uncertainty-for-Trajectory-Prediction}.

\end{abstract}
\section{INTRODUCTION}

\begin{figure}[tb!]
\setlength\abovecaptionskip{3pt}
	\centering  
	 \subfloat[\textbf{Motivation of the Proprioceptive Scenario Gating.}
    ]{
	 	\includegraphics[width=0.52\textwidth, trim=-80 0 -80 0]{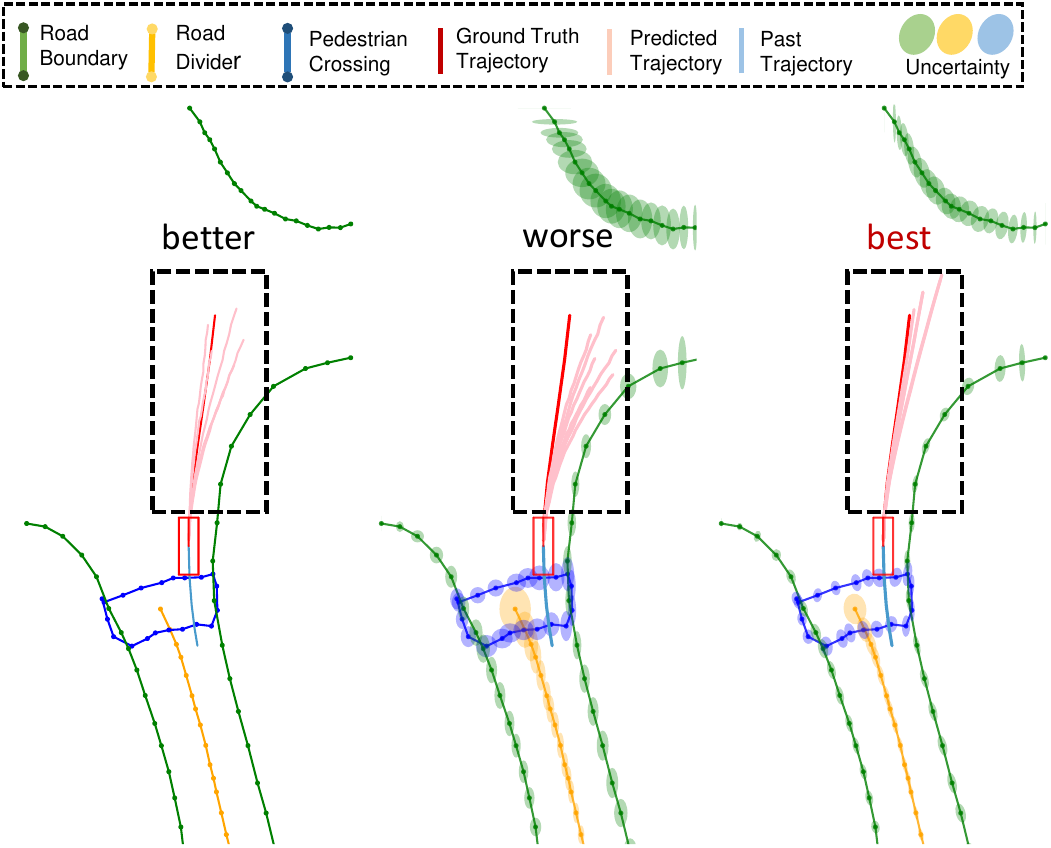}}
     \newline
	\subfloat[\textbf{Better road curvature with the Covariance-Based Uncertainty.}]{
		\includegraphics[width=0.51\textwidth, trim=-80 0 -80 0]{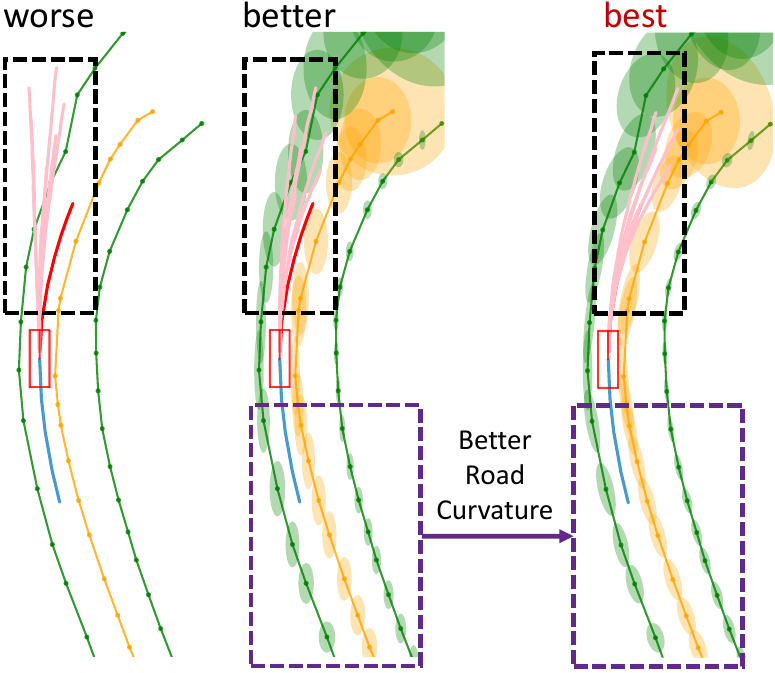}}
        \caption[]{
\small \textbf{Left:} Mapless trajectory prediction baseline (MapTRv2~\cite{maptrv2} + HiVT~\cite{zhou2022hivt}), \textbf{Middle:} Previous uncertainty integration method~\cite{Gu_2024_CVPR}, \textbf{Right:} Ours. In (a), the comparison highlights that predictions enhanced with uncertainty occasionally underperform the baseline, underscoring the necessity of a gating mechanism to selectively incorporate uncertainty. In (b), our Covariance-based Map Uncertainty captures road curvature with good precision that the predicted trajectories align well with the ground truth, rendering the true trajectory invisible.}	
\vspace{-20pt}
\label{fig:teaser}
\vspace{-5pt}
\end{figure}

Traditionally, High-Definition (HD) maps have served as a valuable resource for autonomous driving, supporting tasks in perception~\cite{MapTR, maptrv2, yuan2024streammapnet,li2022hdmapnet,dong2022SuperFusion,liu2022vectormapnet,ding2023pivotnet, zhang2025chameleon}, prediction~\cite{yan2023int2, zheng2024large}, and planning~\cite{zhou2022hivt,GuSunEtAl2021,zhou2023query, ding2024hint}. These maps provide detailed, multi-layered representations of the environment, capturing road geometry (such as polylines and polygons), semantic attributes (e.g., road types), traffic signals, speed limits, and other critical information for static scene understanding. However, creating and maintaining HD maps is both time-consuming and costly, requiring extensive labeling and frequent updates to stay fresh. This dependency on pre-labeled data limits scalability, as HD maps are typically constrained to specific, pre-defined areas, restricting the broader deployment.

 To address these challenges, recent research has shifted towards generating HD maps online directly from sensor data, creating either rasterized bird’s-eye-view (BEV) maps~\cite{philion2020lss,liu2022bevfusion, li2022bevformer} or vectorized representations with polylines~\cite{maptrv2, dong2022SuperFusion, liu2022vectormapnet, ding2023pivotnet}. This ``mapless driving" approach reduces reliance on pre-built, offline maps and minimizes the need for extensive labeling and maintenance. As the field progresses, the focus in mapless driving is evolving from perception (improving online map generation) to motion prediction, with an emphasis on advancing mapless trajectory prediction~\cite{Gu_2024_CVPR, GuSongEtAl2024ECCV}.

In this context, recent studies have highlighted the potential of incorporating uncertainty~\cite{tian2022vibus, Ctrl-u, zhao2020pointly} in online map generation, arising from factors such as occlusions by other vehicles and obstacles, as well as the varying distance to these elements, to enhance mapless trajectory prediction performance \cite{Gu_2024_CVPR}. However, current methods offer limited insight into the specific scenarios where this uncertainty is most beneficial or where it performs poorly (Fig.~\ref{fig:teaser}(a)). In this work, we systematically examine the scenarios in which uncertainty modeling and integration yield the greatest gains in prediction accuracy. Our analysis reveals a key, previously overlooked factor: the agent's kinematic state. Specifically, we find that introducing map uncertainty improves trajectory prediction accuracy primarily in scenarios requiring abrupt changes in vehicle kinematics (e.g., transitioning from a straight to a curved path), while it may degrade performance when historical kinematics are the dominant predictor (e.g., during steady cruising on a straight path). These findings suggest that the effectiveness of online map uncertainty is highly scenario-dependent, prompting us to seek an adaptive approach to selectively apply map uncertainty in trajectory prediction based on scenario-specific conditions.

Specifically, we propose Proprioceptive Scenario Gating, a plug-and-play, lightweight, self-supervised module that adaptively integrates map uncertainty based on the kinematic state of the ego vehicle. By incorporating initial trajectory predictions from a dual parallel stream, this mechanism selectively combines uncertainty-enhanced and non-enhanced results through weighted integration, optimizing final trajectory predictions. Furthermore, aiming to help trajectory prediction methods make better use of road geometry for more accurate predictions, we introduce a Covariance-based Uncertainty Modeling approach to better align uncertainty with map geometry (Fig.~\ref{fig:teaser}(b)). Extensive ablation studies confirm the independent effectiveness of these two techniques, showing consistent performance improvements across various state-of-the-art mapper-predictor configurations.

In summary, our contributions are threefold: \textbf{First}, we systematically demonstrate that the effect of online map uncertainty on mapless trajectory prediction accuracy is scenario-dependent, with its impact varying based on the agent's driving kinematics. \textbf{Second}, we propose a lightweight, self-supervised Proprioceptive Scenario Gating mechanism that adaptively integrates map uncertainty, along with a Covariance-based Map Uncertainty approach that better aligns with map geometry. \textbf{Third}, our approach achieves up to a 23.6\% improvement in trajectory prediction accuracy on the real-world nuScenes benchmark, outperforming existing methods for mapless trajectory prediction.
\section{Related Works}

\subsection{Online Map Estimation}

Online map estimation aims to directly predict the representations of static world elements around an autonomous vehicle from sensor data~\cite{tang2023high}. Early approaches primarily focused on producing 2D BEV rasterized semantic segmentations~\cite{philion2020lss,liu2022bevfusion, can2021stsu,li2022bevformer}. 


Recent advances in vectorized map estimation have focused on improving BEV rasterization by decoding maps into polylines and polygons. Early methods relied on post-processing for map generation~\cite{li2022hdmapnet, dong2022SuperFusion}, while later works introduced end-to-end learning to streamline the process and reduce information loss~\cite{liu2022vectormapnet, ding2023pivotnet}. Studies have also integrated lane perception, inner-instance information, and Euclidean geometry to enhance map representation~\cite{li2023lanesegnet, zhang2023online}.

Further advancements have focused on enhancing localization accuracy and temporal consistency in map representations \cite{yuan2024streammapnet}. Several works have introduced techniques to improve localization and mitigate mapping errors, such as mask-guided instance decoding and anti-disturbance methods for jitter reduction \cite{liu2024mgmap}, as well as hybrid representation learning within end-to-end frameworks \cite{zhou2024himap}. Other studies have explored robustness against sensor corruptions and enhanced point set queries for map construction \cite{liu2025leveraging}.

Despite advancements, the reliability of online-generated maps for downstream tasks remains uncertain. While Gu et al.~\cite{Gu_2024_CVPR} modeled uncertainty using a simple 2 single-variable Laplacian distribution, it limits flexibility for various road geometries. Building on this, we introduce a Covariance-based Map Uncertainty approach that captures more comprehensive uncertainty patterns, enhancing mapping reliability in downstream applications.

\begin{figure}[t!]
\setlength\abovecaptionskip{3pt}
	\centering  
	\subfloat[\textbf{Vehicle's Particle Kinematic Model.} The \textcolor{red}{red} and \textcolor{blue}{blue} lines indicate the future and the past 2s trajectory respectively. $A$ and $B$ denote the first and the last point of the trajectory. $C$ is the center of the vehicle with a heading angle $\Psi$. $O$ and $R$ denote the instantaneous center and the radius of rotation. $\theta_1$ and $\theta_2$ represent the angle between the tangent and the velocity direction at $A$ and $B$ respectively.]{
		\includegraphics[width=0.45\textwidth, trim=-100 0 -100 0]{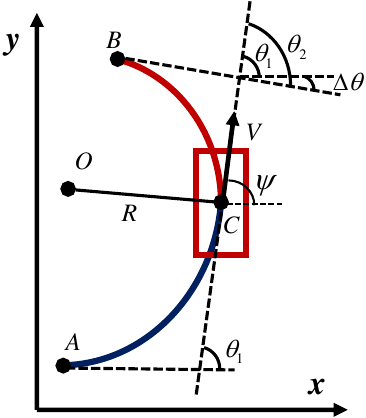}}
    \newline
    \centering
	\subfloat[\textbf{Statistics of the indicator $\Delta \theta$ on nuScenes dataset~\cite{CaesarBankitiEtAl2019} and trajectory prediction performance}. The upper left corner shows the distribution of the dataset across different $\Delta \theta$ intervals. The bar chart shows a comparison of the metrics among the base, previous uncertainty and our methods for each interval. Lower \textcolor{red}{minADE} and \textcolor{blue}{minFDE} values indicate better trajectory prediction accuracy.]{\includegraphics[width=0.45\textwidth, trim=-60 0 -60 0]{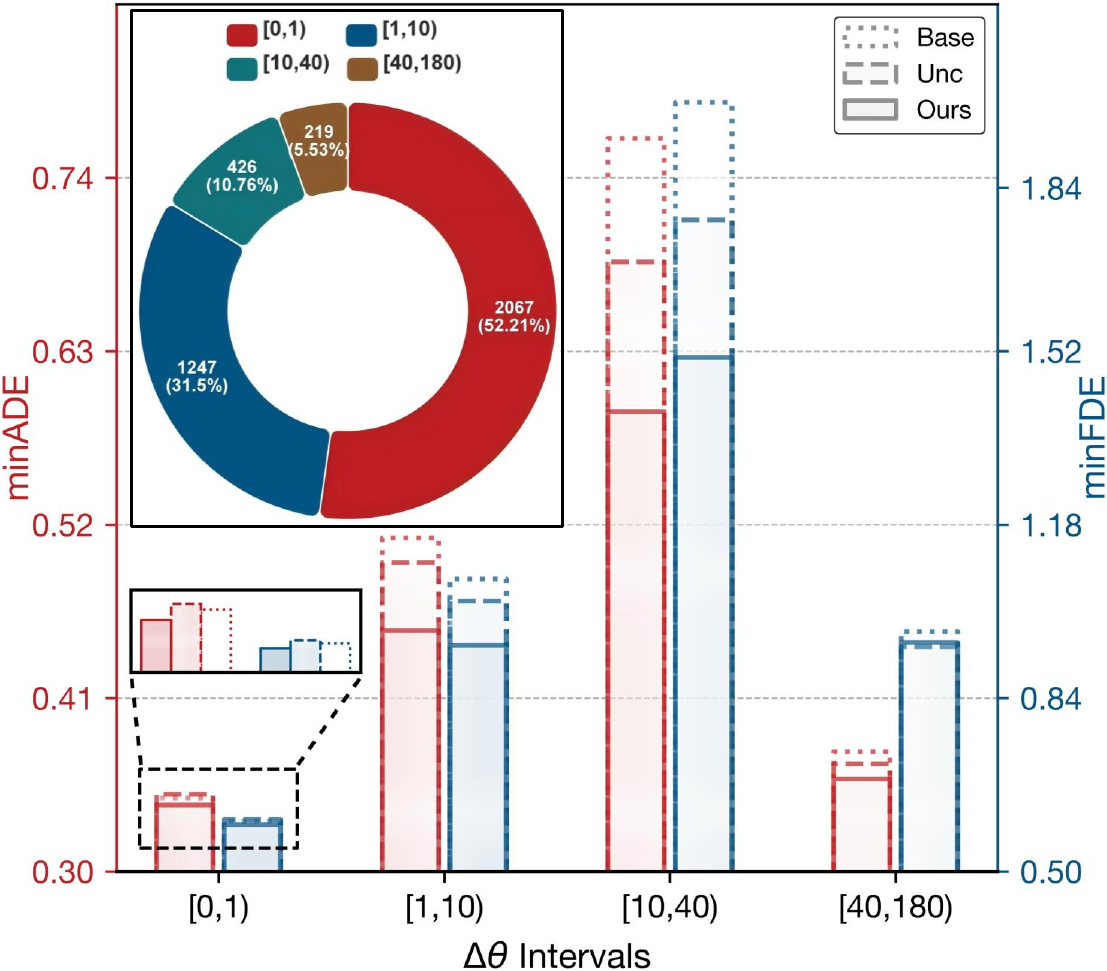}}
    \caption[]{\small \textbf{Preliminary Analysis of Map Uncertainty} for trajectory prediction based ego agent driving kinematic context.\vspace{-2em}}	
 \vspace{-3pt}
        \label{fig:Preliminary Analysis}
\end{figure}

\subsection{Map-Informed Trajectory Prediction}

Trajectory prediction aims to forecast the future paths of agents, based on historical trajectories and scene context. Traditional models often rely on BEV maps encoded with Convolutional Neural Network (CNN) to represent the scene, combining with agent history to inform predictions~\cite{Phan-MinhGrigoreEtAl2020,YuanWengEtAl2021,GillesSabatiniEtAl2021,IvanovicHarrisonEtAl2023}. While effective, these approaches suffer from data inefficiencies and spatial discretization, potentially limiting precision in complex scenes.
To address these limitations, recent work has turned to vectorized map representations, which more efficiently capture detailed map elements relevant to agent motion~\cite{gao2020vectornet,liang2020lanegcn,ZhaoGaoEtAl2020}. Initial methods~\cite{gao2020vectornet, ZhaoGaoEtAl2020, GuSunEtAl2021} introduced Graph Neural Networks (GNNs) to encode lane polylines, capturing map influence on agent behavior. Meanwhile, Transformer-based architectures capture spatial and temporal dependencies, with HiVT~\cite{zhou2022hivt} and QCNet~\cite{zhou2023query} further enhancing efficiency and robustness through query-centric designs and invariant scene representations.

While vectorized approaches have significantly advanced trajectory prediction, challenges remain in effectively integrating them with upstream modules like online mapping. Previous methods encode mapping uncertainty alongside vectorized point sets but lack systematic evaluation of when and where uncertainty should influence prediction~\cite{Gu_2024_CVPR}. Our work addresses these limitations by analyzing the specific conditions under which uncertainty integration is most beneficial and introducing a Proprioceptive Scenario Gating that adapts to varying contexts in real time. This adaptive framework enhances prediction accuracy, robustness, and interpretability in complex driving scenarios.
\section{Method}

This section begins by analyzing where map uncertainty provides benefits for trajectory prediction and where it performs poorly, using the vehicle's particle kinematic model (Sec.~\ref{section3.1}). Based on these insights, we provide an overview of our proposed framework (Sec.~\ref{section3.2}). Then we introduce our Covariance-Based Map Uncertainty (Sec.~\ref{section3.3}) and Proprioceptive Scenario Gating (Sec.~\ref{section3.4}) to improve the current mapless driving paradigm. Finally, we present the overall loss function used to train our model (Sec.~\ref{section3.5}).

\subsection{Analysis of Map Uncertainty for Trajectory Prediction}
\label{section3.1}
We compare the baseline (MapTRv2~\cite{maptrv2} + HiVT~\cite{zhou2022hivt}) and the previous uncertainty method~\cite{Gu_2024_CVPR} for mapless driving to identify scenarios where uncertainty has the most significant impact on trajectory prediction performance.

In Fig.~\ref{fig:Preliminary Analysis}(a), according to the vehicle's particle kinematic model, we have:
\begin{equation}
\theta_i = \dot{\Psi}_i \cdot \Delta t,  i = 1, 2,
\end{equation}
where \(\dot{\Psi}_1\) and \(\dot{\Psi}_2\) denote the average angular velocities of the ego vehicle in the past and future intervals over the same time interval \(\Delta t = 2\) s. Thus, \(\theta_1\) and \(\theta_2\) represent the total rotation experienced by the ego vehicle during the past and future 2 seconds. Then, we have 
\begin{equation}
    \Delta \theta = \left| \theta_1 - \theta_2 \right| = \left|\dot{\Psi}_1  - \dot{\Psi}_2 \right| \cdot \Delta t,
\end{equation}
which quantifies the extent of change \(\left|\dot{\Psi}_1 - \dot{\Psi}_2\right|\) in the ego vehicle's driving state between these two time periods. When $\Delta \theta$ tends to zero, it implies that the vehicle is either traveling in a straight line or making a steady turn.

As shown in Fig.~\ref{fig:Preliminary Analysis}(b), we divide the indicator \(\Delta \theta\) values, which reflects curvature changes and vehicle state variations, into four intervals and analyze the number of scenarios and performance within each. Notably, 52.21\% of the scenarios fall within the range of \([0,1)\), where the baseline (\text{minADE} = \textcolor{red}{0.3466}, \text{minFDE} = \textcolor{blue}{0.5980}) slightly outperforms the uncertainty method (\text{minADE} = \textcolor{red}{0.3490}, \text{minFDE} = \textcolor{blue}{0.6016}), indicating that under consistent driving behavior, excluding uncertainty yields better results. In contrast, with larger future changes in vehicle kinematics, the uncertainty method results in lower prediction errors. As described in the following sections, our proposed method addresses these issues, demonstrating improved performance and achieving better results through experiments.

\begin{figure*}[t]
    \centering
    \includegraphics[width=0.95\textwidth]{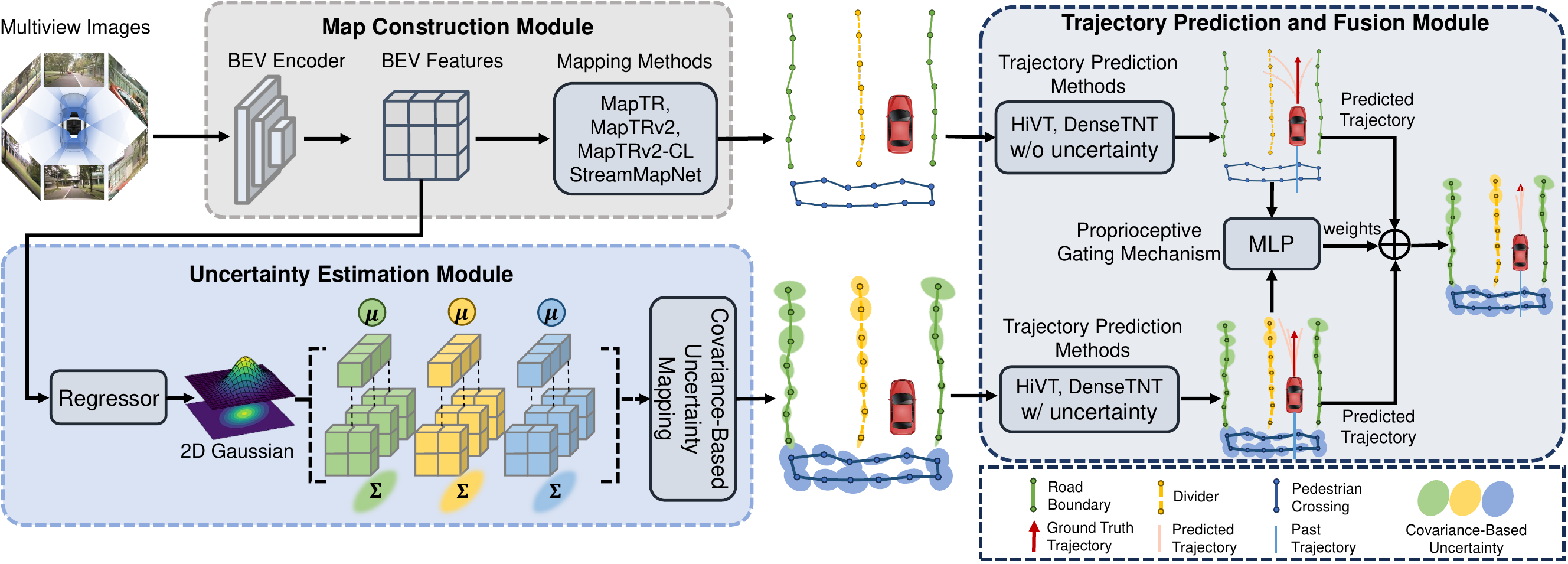}
    \vspace{-8pt}
    \caption{\small \textbf{Model overview.} We first estimate map elements online by encoding multi-view images into a common BEV feature space to regress map element vertices. Each vertex’s uncertainty is modeled using our proposed Covariance-based Uncertainty method, which leverages 2D Gaussian distribution templates. This uncertainty information, along with the original map vertices, is then passed to the downstream trajectory prediction module, which operates in two parallel streams: one that incorporates uncertainty and one that does not. Finally, the proposed Proprioceptive Scenario Gating (MLP network) dynamically adapts the optimal trajectory prediction based on the initial future trajectories prediction from these two streams.}
    \label{fig:overview}
    \vspace{-20pt}
\end{figure*}

\subsection{Architecture Overview}

\label{section3.2}
As shown in Fig.~\ref{fig:overview}, our framework consists of a mapping module based on Covariance Map Uncertainty estimation and a trajectory prediction module with a Proprioceptive Scenario Gating to selectively leverage upstream uncertainty with vehicle kinematics.

First, we use online mapping methods including MapTR~\cite{MapTR}, MapTRv2~\cite{maptrv2}, MapTRv2-Centerline~\cite{maptrv2} and StreamMapNet~\cite{yuan2024streammapnet} to regress map element vertices. For each vertex, uncertainty is modeled with a 2D Gaussian distribution. This uncertainty information is then encoded to refine downstream trajectory predictions. Specifically, we employ two representative models: HiVT~\cite{zhou2022hivt}, which is Transformer-based, and DenseTNT~\cite{GuSunEtAl2021}, which is GNN-based. During their context encoding stage, the map vertex coordinates (mean \(\mu^{(i)}\) in uncertainty modeling) are first encoded via an MLP. For uncertainty modeling, we concatenate the mean \(\mu^{(i)}\), covariance matrix \(\Sigma^{(i)}\) into a single representation, which is then passed through the MLP. This process operates in two parallel streams—one that incorporates uncertainty and one that does not. The Proprioceptive Scenario Gating then dynamically combines the predicted trajectories output from these two streams by passing them through another MLP. This MLP generates weights for each stream, allowing the model to effectively merge the trajectories and produce a more optimal one.

\subsection{Covariance-Based Uncertainty Modeling}
\label{section3.3}

Gu et al.~\cite{Gu_2024_CVPR} models uncertainty in each map element's vertex in the \(x\) and \(y\) directions using 2 single-variable Laplace distribution, assuming independence between \(x\) and \(y\) coordinates. However, in real-world scenarios, these two positional random variables are often correlated due to the directional alignment of the map element. Neglecting these correlations can lead to potentially incorrect modeling of unreliable map regions, which in turn may result in errant trajectory predictions.

To address this, we propose a 2D Gaussian distribution with a correlation coefficient to better represent the spread and orientation of joint uncertainty. Specifically, we use a probabilistic module that jointly predicts both uncertainty parameters (i.e., variance) and the \(x\)-\(y\) correlation coefficient. This extension enables the model to better capture the spatial relationship of uncertainties, offering a more comprehensive representation of map uncertainty that enhances both prediction accuracy and robustness.

Assume a map \( M \) with \( V \) vertices, where each vertex \( v^{(i)} \) in 2D space follows a 2D Gaussian distribution with mean \( \mu^{(i)} \) and covariance \( \Sigma^{(i)} \). The joint probability density function is:
\begin{multline}
f(M | \mu, \Sigma) = \prod_{i=1}^{V} \frac{1}{2\pi |\Sigma^{(i)}|^{1/2}} \exp\left( -\frac{1}{2} (v^{(i)} - \mu^{(i)})^T \right. \\
\quad \left. (\Sigma^{(i)})^{-1} (v^{(i)} - \mu^{(i)}) \right),
\end{multline}
where \( v^{(i)} = \begin{bmatrix} v_1^{(i)} & v_2^{(i)} \end{bmatrix} ^T \) is the coordinate vector of the \( i \)-th vertex, and \( \mu^{(i)} = \begin{bmatrix} \mu_1^{(i)} & \mu_2^{(i)} \end{bmatrix} ^T \) is its mean vector. The covariance matrix for the \( i \)-th vertex is
\(\Sigma^{(i)} = \left[ \sigma_1^2, \rho \sigma_1 \sigma_2; \rho \sigma_1 \sigma_2, \sigma_2^2 \right] \),
where \( \sigma_1 \) and \( \sigma_2 \) are the standard deviations in the x and y dimensions, and \( \rho \) denotes the correlation coefficient, capturing the linear dependency between dimensions.

Compared to modeling uncertainty with two single-variable Laplace distributions in the \(x\) and \(y\) directions (Fig.~\ref{fig:teaser}(b)), our covariance-based method better aligns with the geometry of map elements, such as road edge curvature. We choose a Gaussian distribution with covariance for uncertainty modeling choice as it theoretically outperforms Laplace distribution due to its lighter tails, well-established mathematical, robust properties and reduced sensitivity to outliers.

\subsection{Proprioceptive Scenario Gating}
\label{section3.4}
To address these context-dependent performance discrepancies, we introduce the Proprioceptive Scenario Gating mechanism to enhance trajectory prediction by selectively combining the strengths of both methods. This module enables context-sensitive scenario gating method for more accurate predictions, rather than indiscriminately incorporating uncertainty information.

Traditional exteroceptive gating approaches based on multi-view images or bird’s-eye-view (BEV) data, rely on external sensors (e.g., cameras or LiDAR) to analyze surroundings from multiple perspectives. However, these methods do not consider the car agent inherent properties such as its kinematics. We therefore propose a simple proprioceptive gating method based on the vehicle's internal kinematic data, without relying on any additional environment information, effectively distinguishing scenario types based on trajectory characteristics. As explained in Sec~\ref{section3.2}, the vehicle's trajectories inherently includes its kinematic data, which directly captures position and indirectly incorporates information such as translational velocity, angular velocity, and acceleration, playing a crucial role in distinguishing different scenario types.

Our approach follows a self-supervised paradigm. We employ a dual-encoder structure, where map data with and without uncertainty is processed by a trajectory prediction model to generate predictions \(\mathbf{T}_{\text{base}}\)~$\in~\mathbb{R}^{1 \times 512}$ and \(\mathbf{T}_{\text{unc}}\)~$\in~\mathbb{R}^{1 \times 512}$. These trajectories are flattened, padded and concatenated into an input tensor \(\mathbf{T}\)~$\in~\mathbb{R}^{1 \times 2}$, which is processed through the six-layer MLP network with sizes of 512, 256, 128, 64, 32, 2, reducing the tensor shape from \( [1, 512] \) to \( [1, 2] \):
\begin{equation}
 \mathbf{T} = \text{MLP}(\text{Flatten}( \text{Concat}(\mathbf{T}_{\text{base}}, \mathbf{T}_{\text{unc}}) )) = [ \mathit{v}_{\text{base}}, \mathit{v}_{\text{unc}} ] .  
\end{equation}
The output \([ \mathit{v}_{\text{base}}, \mathit{v}_{\text{unc}} ]\) is transformed via temperature-scaled softmax to obtain weights \([ \mathit{w}_{\text{base}}, \mathit{w}_{\text{unc}} ]\) summing to 1, which serve as labels for training this module.

The gating mechanism processes predicted trajectories to extract and interpret kinematic features relevant to trajectory fusion. This training process enables the model to learn the relationship between kinematic features and optimal trajectory, allowing it to dynamically weight predicted trajectories.

\subsection{Loss Functions}
\label{section3.5}

To train the covariance-based uncertainty model for online mapping, we modify the regression loss to a Negative Log-Likelihood (NLL) loss for a 2D Gaussian distribution:
\begin{multline}
\hspace{-1em}
\mathcal{L}_R (M | \mu, \Sigma) = \sum_{i=1}^{V} \left( \frac{1}{2} \log((2\pi)^2 |\Sigma^{(i)}|) + \frac{1}{2} (v^{(i)} - \mu^{(i)})^T \right. \\
\left. (\Sigma^{(i)})^{-1} (v^{(i)} - \mu^{(i)}) \right),
\end{multline}
where \( \log((2\pi)^2 |\Sigma^{(i)}|) \) is the normalization term related to the determinant of the covariance matrix \( \Sigma^{(i)} \), and \( (v^{(i)} - \mu^{(i)})^T (\Sigma^{(i)})^{-1} (v^{(i)} - \mu^{(i)}) \) represents the error term, measuring the Mahalanobis distance between predicted and actual positions while considering correlations between dimensions. To address instability issues reported in prior work~\cite{Gu_2024_CVPR} with Gaussian distributions during training, we also add a regularization term to prevent gradient explosion, stabilizing the training process and promoting smooth convergence. 

For the Proprioceptive Scenario Gating, it is trained using a Mean Squared Error (MSE) loss function. This loss is computed based on the difference between the module’s predicted scenario weights and the target weights, which reflect the optimal trajectory selection for each scenario. 
\section{Experienments}

\label{sec_exp}
We start with a description of the experimental setup (Sec.\ref{setup}), then compare our method with state-of-the-art map-uncertainty integrated trajectory prediction methods (Sec.\ref{mainresults}). Finally, we provide visualizations (Sec.\ref{visualization}) and an ablation study (Sec.\ref{ablate}).


\begin{table*}
  \centering
  \setlength{\tabcolsep}{3pt}
  \resizebox{1\linewidth}{!}{
  \begin{tabular}{l|llll|llll}
    \toprule
    Prediction Method & \multicolumn{4}{c|}{HiVT~\cite{zhou2022hivt}} & \multicolumn{4}{c}{DenseTNT~\cite{GuSunEtAl2021}} \\
    \midrule
    Online HD Map Method & minADE $\downarrow$ & minFDE $\downarrow$ & MR $\downarrow$ &FPS & minADE $\downarrow$ & minFDE $\downarrow$ & MR $\downarrow$ &FPS\\
    \midrule
    MapTR~\cite{MapTR} & 0.4015 & 0.8404 & 0.0960 &17.91 & 1.1228 & 2.2151 & 0.3726 &7.36\\
    MapTR~\cite{MapTR} + Unc~\cite{Gu_2024_CVPR} & 0.3910  & 0.8049  & 0.0818 &16.65 & 1.1946  & 2.2666  & 0.3848 &7.75 \\
    \rowcolor{gray!20} 
    MapTR~\cite{MapTR} + Ours & \textbf{0.3672} {\small \color{darkpastelgreen} ($\mathbf{-6.1}\%$)} & \textbf{0.7395} {\small \color{darkpastelgreen} ($\mathbf{-8.1}\%$)} & \textbf{0.0756} {\small \color{darkpastelgreen} ($\mathbf{-7.6}\%$)}&15.08 & \textbf{1.0865} {\small \color{darkpastelgreen} ($\mathbf{-9.0}\%$)} & \textbf{2.0969} {\small \color{darkpastelgreen} ($\mathbf{-7.5}\%$)} & \textbf{0.3728} {\small \color{darkpastelgreen} ($\mathbf{-3.2}\%$)} &6.96\\
    \midrule
    MapTRv2~\cite{maptrv2} & 0.4017 & 0.8406 & 0.0959& 15.66 & 1.3262 & 2.5687 & 0.4301 &7.57 \\ 
    MapTRv2~\cite{maptrv2} + Unc~\cite{Gu_2024_CVPR} & 0.3913 & 0.8054 & 0.0819 &14.98 & 1.3256 & 2.6390 & 0.4435 &7.38\\
    \rowcolor{gray!20} 
    MapTRv2~\cite{maptrv2} + Ours & \textbf{0.3670} {\small \color{darkpastelgreen} ($\mathbf{-6.2}\%$)}& \textbf{0.7538} {\small \color{darkpastelgreen} ($\mathbf{-6.4}\%$)}&\textbf{0.0708} {\small \color{darkpastelgreen} ($\mathbf{-13.6}\%$)} &13.65& \textbf{1.1585} {\small \color{darkpastelgreen} ($\mathbf{-12.6}\%$)}& \textbf{2.4566} {\small \color{darkpastelgreen} ($\mathbf{-7.4}\%$)}& \textbf{0.3891} {\small \color{darkpastelgreen} ($\mathbf{-12.3}\%$)} &6.89\\
    \midrule
    MapTRv2-CL~\cite{maptrv2} &0.3789 & 0.7859 & 0.0865 & 14.02 & 0.8333 & 1.4752 & 0.1719 &7.62 \\
    MapTRv2-CL~\cite{maptrv2} + Unc~\cite{Gu_2024_CVPR} &0.3674 & 0.7418 & 0.0739 &14.03 & 0.9666 & 1.6439 & 0.2082 &7.29\\
    \rowcolor{gray!20} 
    MapTRv2-CL~\cite{maptrv2} + Ours & \textbf{0.3659} {\small \color{darkpastelgreen} ($\mathbf{-0.4}\%$)}& \textbf{0.7404} {\small \color{darkpastelgreen} ($\mathbf{-0.2}\%$)}& \textbf{0.0721} {\small \color{darkpastelgreen} ($\mathbf{-2.4}\%$)} &13.47 & \textbf{0.7787} {\small \color{darkpastelgreen} ($\mathbf{-19.4}\%$)}& \textbf{1.4662} {\small \color{darkpastelgreen} ($\mathbf{-10.8}\%$)}& \textbf{0.1590} {\small \color{darkpastelgreen} ($\mathbf{-23.6}\%$)} &6.73\\
    \midrule
    StreamMapNet~\cite{yuan2024streammapnet}  & 0.3963 & 0.8223 & 0.0923 & 14.17 & 1.0639 & 2.1430 & 0.3412 &7.42\\
    StreamMapNet~\cite{yuan2024streammapnet}  + Unc~\cite{Gu_2024_CVPR} & 0.3899 & 0.8101 & 0.0861 &13.69 & 1.0902 & 2.1412 & 0.3261 &6.91\\
    \rowcolor{gray!20} 
    StreamMapNet~\cite{yuan2024streammapnet} + Ours & \textbf{0.3828} {\small \color{darkpastelgreen} ($\mathbf{-1.8}\%$)}& \textbf{0.7981} {\small \color{darkpastelgreen} ($\mathbf{-1.5}\%$)}& \textbf{0.0834} {\small \color{darkpastelgreen} ($\mathbf{-0.3}\%$)} &13.26 & \textbf{0.9675} {\small \color{darkpastelgreen} ($\mathbf{-11.3}\%$)}& \textbf{1.6883} {\small \color{darkpastelgreen} ($\mathbf{-21.2}\%$)}& \textbf{0.2628} {\small \color{darkpastelgreen} ($\mathbf{-19.4}\%$)} &6.32\\
    \bottomrule
  \end{tabular}
  }
  \caption{\small Quantitative results of trajectory prediction for representative mapping/prediction model combinations on the nuScenes~\cite{CaesarBankitiEtAl2019} dataset under different settings. Our method incorporating the proposed Proprioceptive Scenario Gating and Covariance-based Map Uncertainty consistently improves prediction model performance, surpassing previous uncertainty integration methods across nearly all combinations. Additionally, we benchmark FPS on an NVIDIA RTX 3090, demonstrating that our method achieves comparable computational efficiency to the baseline while delivering superior predictive performance.}
  \label{tab:main}
  \vspace{-10pt}
\end{table*}

\subsection{Experiment Setup}
\label{setup}

\textbf{Dataset.} We evaluate our method on the nuScenes dataset~\cite{CaesarBankitiEtAl2019} which is divided into 500 for training, 200 for validation, and 150 for testing. Each 20-second scene features keyframe annotations at 2 Hz, with six multi-view RGB cameras providing a 360° view around the ego-vehicle. We use the same trajectory processing method as in~\cite{Gu_2024_CVPR}. Each model then forecasts motion 3 seconds into the future based on 2 seconds of historical data.

\textbf{Metrics.}
For trajectory prediction evaluation, we adopt standard metrics commonly used in recent prediction challenges~\cite{ChangLambertEtAl2019,wilson2021argoverse2,sun2020scalability}: minimum Average Displacement Error (minADE), minimum Final Displacement Error (minFDE), and Miss Rate (MR). Our model outputs six candidate trajectories per agent. The minADE metric computes the average Euclidean ($\ell_2$) distance between the most accurate predicted trajectory and the ground truth (GT) trajectory across all time steps, while minFDE measures distance at the final time step. MR quantifies the percentage of predictions where the endpoint of the best trajectory deviates from the GT endpoint by more than 2 meters.

\textbf{Implementation Details.}
All models are trained on a single NVIDIA GeForce RTX 3090 GPU. For each mapping model, the learning rate is set to 1.5e-4, the regression loss weight is 0.03, and gradient norms are clipped at 3.

For the trajectory prediction task, HiVT is trained with a learning rate of 5.0e-4, a batch size of 32 and a temprature of 0.6. For DenseTNT, the learning rate remains at 5.0e-4, while the batch size and temperature are adjusted to 16 and 0.5, respectively. A dropout rate of 0.1 is applied consistently across all trajectory prediction models to prevent overfitting.

\subsection{Main Results}
\label{mainresults}

To evaluate the impact of our uncertainty modeling strategy on downstream trajectory prediction, we conduct experiments across multiple combinations of online map estimation and prediction methods. Aiming to facilitate a more effective comparison with previous work, we select the same four mapping methods (MapTR~\cite{MapTR}, MapTRv2~\cite{maptrv2}, MapTRv2-Centerline~\cite{maptrv2} and StreamMapNet~\cite{yuan2024streammapnet}
) and two trajectory prediction models (HiVT~\cite{zhou2022hivt} and DenseTNT~\cite{GuSunEtAl2021}), resulting in eight combinations. Table.~\ref{tab:main} shows consistent performance improvements across all combinations when using our formulation. With HiVT, our method improves minADE and minFDE by over 6\% on MapTR and MapTRv2, and achieves a notable 13.6\% improvement in MR on MapTRv2. 
For DenseTNT, the largest gains are observed on MapTRv2-Centerline, where minADE, minFDE, and MR improve by 19.4\%, 10.8\%, and 23.6\%, respectively. Fig.~\ref{fig:Preliminary Analysis}(b) further illustrates that across different intervals of \( \Delta \theta \), our method consistently improves trajectory prediction performance.

\begin{figure*}[t]
    \centering
    \includegraphics[width=1.0\textwidth, trim=10 20 10 0 ]{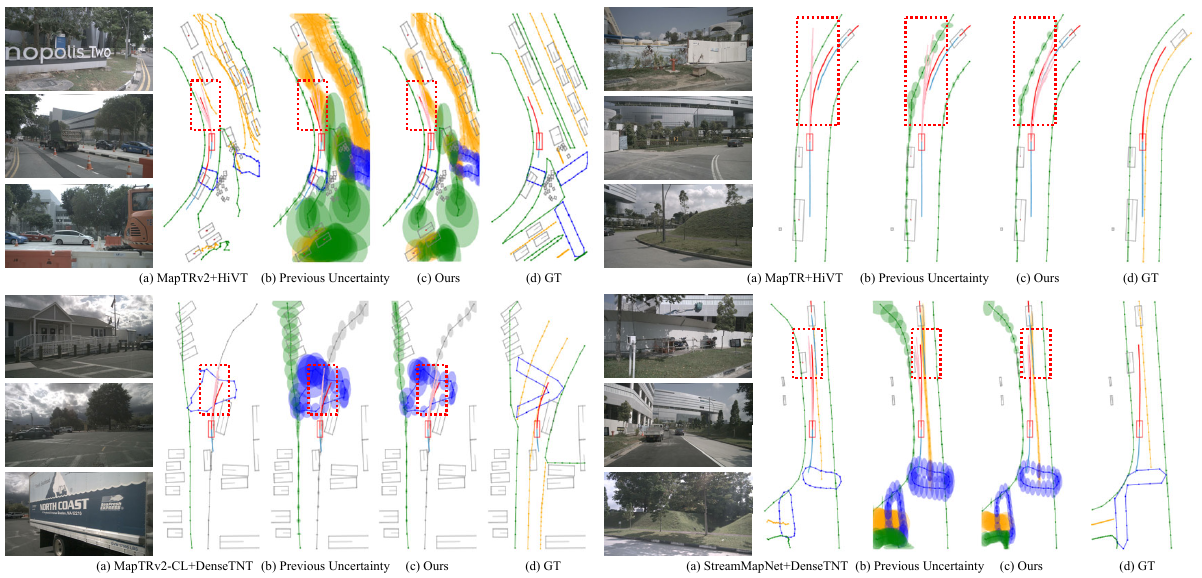}
    \vspace{1em}
    \caption{\small Qualitative visualization showing various combinations of online mapping and trajectory prediction models. Our approach outperforms both the baseline and the state-of-the-art uncertainty modeling method~\cite{Gu_2024_CVPR} in all these driving scenarios.} 
    \vspace{-20pt}
    \label{fig:vis-combine}
\end{figure*}

\subsection{Visualizations}
\label{visualization}
Qualitative results in Fig.~\ref{fig:vis-combine} demonstrate that our method consistently improves trajectory prediction across various driving scenarios. In the \textbf{top left} of Fig.~\ref{fig:vis-combine}, the ego vehicle executes a steady circular turn through a complex scenario, where \(\Delta \theta \approx 0\) indicates a stable driving state (Sec.~\ref{section3.2}). The baseline (a) outperforms the previous uncertainty-based method (b), yet part of the predicted trajectory extends beyond the road boundary. In contrast, our method (c) aligns closely with the ground truth trajectory, remaining within the drivable area. In the \textbf{top right}, as the ego vehicle approaches a gentle curve, the previous method (b) marginally outperforms the baseline (a), but both predictions exceed the road boundary. Our method (c) follows the road curvature and avoids the vehicle ahead. The \textbf{Bottom left} visualizes the ego vehicle avoiding a truck while staying in its lane. Both the baseline and previous uncertainty-based method drift into the opposite lane, while our method successfully avoids the truck and maintains on its current lane. In the \textbf{bottom right}, the ego vehicle transitions from a turn to a straight path. The previous method underperforms compared to the baseline, while our method outperforms both, demonstrating the advantages of our gating mechanism in leveraging kinematic information. As shown in Table~\ref{tab:main}, for DenseTNT, previous uncertainty-based methods generally underperform compared to the baseline. The \textbf{bottom-left} and \textbf{bottom-right} scenarios in Fig.~\ref{fig:vis-combine} highlight that our method produces more precise and tightly clustered trajectories.

\subsection{Ablations}
\label{ablate}

\subsubsection{Covariance-based Map Uncertainty}
First, we evaluate the effectiveness of the proposed covariance-based uncertainty without using the Proprioceptive Scenario Gating, as shown in Table~\ref{tab:ablation-covariance}. This formulation consistently enhances trajectory prediction performance across all mapping methods, offering richer, geometry-aligned uncertainty information to downstream tasks. The inclusion of covariance-based uncertainty significantly reduces minFDE and MR, with the most notable improvements observed in MapTRv2-Centerline (4.3\% reduction in minFDE) and StreamMapNet (6.5\% reduction in MR). We also explore the impact of modeling uncertainty with four different distributions, as shown in Table~\ref{tab:ablation-covariance-2}. Our results indicate that whether using Laplace~\cite{Gu_2024_CVPR} or Gaussian distributions, incorporating covariance (correlation) into uncertainty modeling consistently outperforms the independent distribution approach, as it better aligns with road curvature. Moreover, covariance-based Gaussian distribution generally outperforms the covariance-based Laplace distribution. This further validates our choice of using joint Gaussian distribution for modeling uncertainty.

\vspace{-0.1cm}

\begin{table}
  \centering
  \resizebox{1\linewidth}{!}{
  \begin{tabular}{l|lll}
    \toprule
    Prediction Method & \multicolumn{3}{c}{HiVT~\cite{zhou2022hivt}} \\
    \midrule
    Online HD Map Method & minADE $\downarrow$ & minFDE $\downarrow$ & MR $\downarrow$\\
    \midrule
    MapTR~\cite{MapTR} & 0.4015 & 0.8404 & 0.0960 \\
    MapTR~\cite{MapTR} + Unc~\cite{Gu_2024_CVPR} & 0.3910 & 0.8049  & 0.0818 \\
    \rowcolor{gray!20} 
    MapTR~\cite{MapTR} + CovMat 
& \textbf{0.3851} {\small \color{darkpastelgreen}($\mathbf{-1.5}\%$)}  
& \textbf{0.7812} {\small \color{darkpastelgreen}($\mathbf{-2.9}\%$)}   
& \textbf{0.0815} {\small \color{darkpastelgreen}($\mathbf{-0.4}\%$)}  \\
    
    \midrule
    MapTRv2~\cite{maptrv2} & 0.4017 & 0.8406 & 0.0959\\ 
    MapTRv2~\cite{maptrv2} + Unc~\cite{Gu_2024_CVPR} & 0.3913 & 0.8054 & 0.0819\\
    \rowcolor{gray!20} 
    MapTRv2~\cite{maptrv2} + CovMat 
& \textbf{0.3776} {\small \color{darkpastelgreen}($\mathbf{-3.5}\%$)} 
& \textbf{0.7707} {\small \color{darkpastelgreen}($\mathbf{-4.3}\%$)}
& \textbf{0.0789} {\small \color{darkpastelgreen}($\mathbf{-1.5}\%$)}\\

    \midrule
    MapTRv2-CL~\cite{maptrv2} & 0.3789 & 0.7859 & 0.0865\\
    MapTRv2-CL~\cite{maptrv2} + Unc~\cite{Gu_2024_CVPR} & 0.3674 & 0.7418 & 0.0739\\
    \rowcolor{gray!20} 
    MapTRv2-CL~\cite{maptrv2} + CovMat 
& \textbf{0.3662} {\small \color{darkpastelgreen}($\mathbf{-0.3}\%$)}
& \textbf{0.7411} {\small \color{darkpastelgreen}($\mathbf{-0.1}\%$)}
& \textbf{0.0728} {\small \color{darkpastelgreen}($\mathbf{-1.5}\%$)}\\

    \midrule
    StreamMapNet~\cite{yuan2024streammapnet}  & 0.3963 & 0.8223 & 0.0923\\
    StreamMapNet~\cite{yuan2024streammapnet}  + Unc~\cite{Gu_2024_CVPR} & 0.3899 & 0.8101 & 0.0861 \\
    \rowcolor{gray!20} 
    StreamMapNet~\cite{yuan2024streammapnet}  + CovMat 
& \textbf{0.3843} {\small \color{darkpastelgreen}($\mathbf{-1.4}\%$)}
& \textbf{0.8035} {\small \color{darkpastelgreen}($\mathbf{-0.8}\%$)}
& \textbf{0.0805} {\small \color{darkpastelgreen}($\mathbf{-6.5}\%$)}\\
    \bottomrule
  \end{tabular}
  }
    \caption{\small In general, leveraging our covariance-based uncertainty formulation (CovMat) for online mapping consistently improves prediction model performance across all combinations compared to the previous two single-variable Laplacian approach~\cite{Gu_2024_CVPR}.}
  \label{tab:ablation-covariance}

\end{table}

\begin{table}
  \centering
  \resizebox{1\linewidth}{!}{
  \begin{tabular}{l|lll}
    \toprule
    Prediction Method & \multicolumn{3}{c}{ HiVT~\cite{zhou2022hivt}} \\
    \midrule
    Online HD Map Method & minADE $\downarrow$ & minFDE $\downarrow$ & MR $\downarrow$\\
    \midrule
    MapTR~\cite{MapTR} & 0.4015 & 0.8404 & 0.0960 \\
    MapTR~\cite{MapTR} + Bivariate Laplace w/o Covariance~\cite{Gu_2024_CVPR} & 0.3910 & 0.8049  & 0.0802 \\
     MapTR~\cite{MapTR} + Bivariate  Gaussian w/o Covariance& 0.3931 & 0.8079  & 0.0839 \\
      MapTR~\cite{MapTR} + Bivariate  Laplace w/ Covariance& 0.3853 & 0.7862  & \textbf{0.0803} \\
    \rowcolor{gray!20} 
    MapTR~\cite{MapTR} + Bivariate Gaussian w/ Covariance (Ours) 
& \textbf{0.3851}   
& \textbf{0.7812}    
& 0.0815   \\

    \bottomrule
  \end{tabular}
  }
    \caption{\small Investigating the impact of modeling uncertainty using four different probability distributions on trajectory prediction. The results show that the covariance-based joint Gaussian distribution provides the best modeling approach.}
  \label{tab:ablation-covariance-2}
  \vspace{-20pt}
\end{table}

\subsubsection{Proprioceptive Scenario Gating}
We ablate the effectiveness of the proposed Proprioceptive Scenario Gating. Table~\ref{tab:ablation-scenario-Gating-1} shows that adaptively selecting the optimal trajectory based on kinematic features improves prediction accuracy across all mapping-prediction combinations. By analyzing the values of \(\mathit{w}_{\text{base}}\) and \(\mathit{w}_{\text{base}}\), we find that in stable vehicle states ( \(\Delta\theta\) ~$\in~[0,1)$), the average of \(\mathit{w}_{\text{base}}\) is 0.82, while in other cases, \(\mathit{w}_{\text{base}}\) has an average value of 0.37. This further demonstrates that our gating mechanism learns vehicle kinematic information from the trajectory and makes a more optimal trajectory.

\begin{table}
\vspace{-0.6cm}

  \centering
  \resizebox{1\linewidth}{!}{
  \begin{tabular}{l|lll}
    \toprule
    Prediction Method & \multicolumn{3}{c}{HiVT~\cite{zhou2022hivt}} \\
    \midrule
    Online HD Map Method & minADE $\downarrow$ & minFDE $\downarrow$ & MR $\downarrow$ \\
    \midrule
    MapTR~\cite{MapTR} & 0.4015 & 0.8404 & 0.0960  \\
    MapTR~\cite{MapTR} + Unc~\cite{Gu_2024_CVPR} & 0.3910 & 0.8049  & 0.0818  \\
    \rowcolor{gray!20} 
    MapTR~\cite{MapTR} + Gating 
& \textbf{0.3829}  {\small \color{darkpastelgreen} ($\mathbf{-2.1}\%$)}
& \textbf{0.7855} {\small \color{darkpastelgreen} ($\mathbf{-2.4}\%$)} 
& \textbf{0.0805} {\small \color{darkpastelgreen} ($\mathbf{-1.6}\%$)}  \\
    \midrule
    MapTRv2~\cite{maptrv2} & 0.4017 & 0.8406 & 0.0959 \\ 
    MapTRv2~\cite{maptrv2} + Unc~\cite{Gu_2024_CVPR} & 0.3913 & 0.8054 & 0.0819 \\
    \rowcolor{gray!20} 
    MapTRv2~\cite{maptrv2} + Gating 
& \textbf{0.3815} {\small \color{darkpastelgreen} ($\mathbf{-2.5}\%$)} 
& \textbf{0.7914} {\small \color{darkpastelgreen} ($\mathbf{-1.7}\%$)}
& \textbf{0.0807} {\small \color{darkpastelgreen} ($\mathbf{-1.5}\%$)}  \\
    \midrule
    MapTRv2-CL~\cite{maptrv2} & 0.3789 & 0.7859 & 0.0865\\
    MapTRv2-CL~\cite{maptrv2} + Unc~\cite{Gu_2024_CVPR} & 0.3674 & 0.7418 & 0.0739 \\
    \rowcolor{gray!20} 
    MapTRv2-CL~\cite{maptrv2} + Gating 
& \textbf{0.3650} {\small \color{darkpastelgreen} ($\mathbf{-0.7}\%$)}
& \textbf{0.7362} {\small \color{darkpastelgreen} ($\mathbf{-0.8}\%$)}
& \textbf{0.0685} {\small \color{darkpastelgreen} ($\mathbf{-7.3}\%$)} \\
    \midrule
    StreamMapNet~\cite{yuan2024streammapnet}  & 0.3963 & 0.8223 & 0.0923 \\
    StreamMapNet~\cite{yuan2024streammapnet}  + Unc~\cite{Gu_2024_CVPR} & 0.3899 & 0.8101 & 0.0861 \\
    \rowcolor{gray!20} 
    StreamMapNet~\cite{yuan2024streammapnet} + Gating 
& \textbf{0.3858} {\small \color{darkpastelgreen}($\mathbf{-1.0}\%$)}  
& \textbf{0.7981}  {\small \color{darkpastelgreen}($\mathbf{-1.5}\%$)} 
& \textbf{0.0847} {\small \color{darkpastelgreen}($\mathbf{-1.6}\%$)}    \\
    \bottomrule
  \end{tabular}
  }
    \caption{\small Incorporating  Gating method to dynamically leverage map uncertainty improves trajectory prediction consistently.\vspace{0em}}
  \label{tab:ablation-scenario-Gating-1}
  \vspace{-12pt}
\end{table}

\subsubsection{Dynamic Proprioception vs. Exteroceptive Sensing}
To validate the effectiveness of our dynamic proprioception-based approach, we compare it against exteroceptive sensing methods using the CLIP model~\cite{radford2021learning} and ResNet50~\cite{he2016deep}. Specifically, front camera images are inputs of the CLIP image encoder or ResNet50 to extract semantic features, which are then processed by a multi-layer perceptron to generate a weighted trajectory selection based on classification results. The road semantics are divided into four categories as ground truth: labels — \textit{low-curvature turn}, \textit{high-curvature turn}, \textit{presence of other agents}, and \textit{straight path} — each represented by one-hot encodings.


As shown in Table~\ref{tab:ablation-scenario-Gating-CLIP}, the CLIP-based and ResNet-based exteroceptive approaches underperform our dynamic proprioception-based method across all models and metrics. In some cases, it even falls below baseline performance, suggesting that trajectory selection optimization based solely on semantic scene understanding may be insufficient. Additionally, Table~\ref{tab: Inference Time Comparison} highlights a substantial difference in FPS, with our method achieving speeds up to 30 times higher than CLIP and 10 times higher than ResNet50, making it more suitable for real-time deployment in vehicles.
\vspace{-5pt}

\begin{table}
  \centering
  \resizebox{1\linewidth}{!}{
  \begin{tabular}{l|lll}
    \toprule
    Prediction Method & \multicolumn{3}{c}{HiVT~\cite{zhou2022hivt}}\\
    \midrule
    Online HD Map Method & minADE $\downarrow$ & minFDE $\downarrow$ & MR $\downarrow$\\
    \midrule
    MapTR~\cite{MapTR} & 0.4015 & 0.8404 & 0.0960\\
    MapTR~\cite{MapTR} + CLIP~\cite{radford2021learning} & 0.4200 & 0.8821 & 0.1040\\
    MapTR~\cite{MapTR} + ResNet50~\cite{he2016deep} & 0.4248 & 0.9008 & 0.1060\\
    \rowcolor{gray!20} 
    MapTR~\cite{MapTR} + Ours 
& \textbf{0.3672} 
& \textbf{0.7395} 
& \textbf{0.0756}\\ 
    \midrule
    MapTRv2~\cite{maptrv2} & 0.4017 & 0.8406 & 0.0959\\ 
    MapTRv2~\cite{maptrv2} + CLIP~\cite{radford2021learning}~ & 0.3988 & 0.8370 & 0.0969\\
    MapTRv2~\cite{maptrv2} + ResNet50~\cite{he2016deep} & 0.4191 & 0.8751 & 0.1041\\
    \rowcolor{gray!20} 
    MapTRv2~\cite{maptrv2} + Ours 
& \textbf{0.3670} 
& \textbf{0.7538} 
& \textbf{0.0708} \\
    \midrule
    MapTRv2-CL~\cite{maptrv2} & 0.3789 & 0.7859 & 0.0865 \\
    MapTRv2-CL~\cite{maptrv2} + CLIP~\cite{radford2021learning} & 0.4433 & 0.9711 & 0.1124 \\
    MapTRv2-CL~\cite{maptrv2} + ResNet50~\cite{he2016deep} & 0.4211 & 0.9058 & 0.1052\\
    \rowcolor{gray!20} 
    MapTRv2-CL~\cite{maptrv2} + Ours 
& \textbf{0.3659} 
& \textbf{0.7404} 
& \textbf{0.0721}\\ 
    \midrule
    StreamMapNet~\cite{yuan2024streammapnet}  & 0.3963 & 0.8223 & 0.0923 \\
    StreamMapNet~\cite{yuan2024streammapnet}  + CLIP~\cite{radford2021learning} & 0.4191 & 0.9010 & 0.0926 \\
    StreamMapNet~\cite{yuan2024streammapnet} + ResNet50~\cite{he2016deep} & 0.4199 & 0.9024 & 0.0988\\
    \rowcolor{gray!20} 
    StreamMapNet~\cite{yuan2024streammapnet} + Ours 
&\textbf{0.3828} 
&\textbf{0.7981} 
&\textbf{0.0834}\\ 
    \bottomrule
  \end{tabular}
  }
    \caption{\small Our Proprioceptive Scenario Gating outperforms the CLIP-based and ResNet-based exteroceptive alternatives in prediction accuracy across all mapping-prediction combinations.}
  \label{tab:ablation-scenario-Gating-CLIP}

\end{table}


\begin{table}
  \centering
  \resizebox{1\linewidth}{!}{
  \begin{tabular}{l|l|l|>{\columncolor{gray!20}}l}
    \toprule
   Gating Method & \multicolumn{1}{c|}{CLIP~\cite{radford2021learning}} &\multicolumn{1}{c|}{ResNet50~\cite{he2016deep}}
    & \multicolumn{1}{>{\columncolor{gray!20}}c}{Ours} \\
    \midrule
    Online HD Map Method & FPS & FPS & FPS  \\
    \midrule
    MapTR~\cite{MapTR} & 2.21 & 6.59& \textbf{52.26} \\
    MapTRv2~\cite{maptrv2} & 2.15 & 6.51 & \textbf{92.09} \\ 
    MapTRv2-CL~\cite{maptrv2} & 2.33 & 6.70 & \textbf{87.35} \\
    StreamMapNet~\cite{yuan2024streammapnet}  & 2.42 & 6.91 & \textbf{84.57}  \\
    \bottomrule
  \end{tabular}
  }
     \caption{\small Our Proprioceptive Scenario Gating achieves FPS up to 30 times higher than the CLIP-based exteroceptive alternative approach and 10 times higher than the ResNet50-based exteroceptive method.}
  \label{tab: Inference Time Comparison}

\end{table}

\subsection{Scalability}

To validate the scalability of our proposed architecture, we replace the baseline (MapTR~\cite{MapTR}) in the dual-stream structure with the BEV feature attention method~\cite{GuSongEtAl2024ECCV}, keeping the rest of the architecture unchanged. The results, as shown in Table~\ref{tab:scalibility}, demonstrate that our approach achieves SOTA performance in trajectory prediction tasks, highlighting the strong generalization capability of our architecture.
\begin{table}
\vspace{-0.8cm}
  \centering
  \resizebox{1\linewidth}{!}{
  \begin{tabular}{l|lll}
    \toprule
    Prediction Method & \multicolumn{3}{c}{HiVT~\cite{zhou2022hivt}} \\
    \midrule
    Online HD Map Method & minADE $\downarrow$ & minFDE $\downarrow$ & MR $\downarrow$ \\
    \midrule
    MapTR~\cite{MapTR} + Unc~\cite{Gu_2024_CVPR} & 0.3910 & 0.8049 & 0.0818  \\
    MapTR~\cite{MapTR} + BEV Feature Attention~\cite{GuSongEtAl2024ECCV} & 0.3617 & 0.7401 & 0.0720  \\
    MapTR~\cite{MapTR} + CovMat & 0.3851 & 0.7812  & 0.0815  \\
    \rowcolor{gray!20} 
    MapTR~\cite{MapTR} + Ours 
& \textbf{0.3494}  {\small \color{darkpastelgreen} ($\mathbf{-3.5}\%$)}
& \textbf{0.7028} {\small \color{darkpastelgreen} ($\mathbf{-5.6}\%$)} 
& \textbf{0.0653} {\small \color{darkpastelgreen} ($\mathbf{-9.3}\%$)}  \\
    \bottomrule
  \end{tabular}
  }
    \caption{\small Repalcing the baseline (MapTR~\cite{MapTR}) in our proposed dual-stream structure with a stronger alternative~\cite{GuSongEtAl2024ECCV}, showing significant improvements in prediction accuracy.}
  \label{tab:scalibility}

\end{table}
\newpage
\section{Conclusion}


In this work, we explore the impact of map uncertainty for mapless trajectory prediction and identify the critical role of the vehicle’s kinematic state in determining when such uncertainty is beneficial. Building on these insights, we introduce Proprioceptive Scenario Gating, a novel mechanism adaptively integrating map uncertainty into trajectory prediction based on forecasts of the vehicle’s future kinematics. This self-supervised approach not only improves performance but also enhances interpretability by identifying scenarios where uncertainty contributes positively. Furthermore, we propose a Covariance-based Map Uncertainty modeling that better aligns with map geometry, leading to further performance improvements. Extensive experiments on the real-world data demonstrate the significant performance gains of our method. 
\vspace{-10pt}
{
    \small
    \bibliographystyle{ieeetr}
    \bibliography{main}
}

\end{document}